\RequirePackage{fix-cm}
\documentclass[twocolumn]{svjour3}          
\smartqed  
\usepackage{graphicx}
\usepackage{natbib}
\usepackage{multirow}
\usepackage{amsmath}
\usepackage{amssymb}
\usepackage{bbm}
\usepackage{bm}
\usepackage{booktabs}
\usepackage{array} 
\usepackage{blindtext}
\usepackage{comment}

\usepackage{pifont}

\usepackage{xspace}
\newcommand{\name}{MotionDiffuse\xspace}
\newcommand{\eg}{\textit{e.g.}\xspace}

\usepackage[dvipsnames]{xcolor}
\usepackage{color, colortbl}
\definecolor{citecolor}{HTML}{0071bc}
\definecolor{tabhighlight}{HTML}{e5e5e5}
\usepackage[colorlinks,citecolor=citecolor]{hyperref}

\makeatletter
\renewcommand\paragraph{
  \@startsection{paragraph} 
  {4} 
  {\z@} 
  {.5em \@plus1ex \@minus.2ex} 
  {-.5em} 
  {\normalfont\normalsize\bfseries} 
}
\makeatother
\begin{document}
\sloppy

\title{MotionDiffuse: Text-Driven Human Motion Generation with Diffusion Model
}


\author{Mingyuan Zhang* \and
        Zhongang Cai* \and
        Liang Pan \and
        Fangzhou Hong \and
        Xinying Guo \and
        Lei Yang \and
        Ziwei Liu
}

\authorrunning{Mingyuan Zhang, Zhongang Cai et al.} 

\institute{Mingyuan Zhang \at
              S-Lab, Nanyang Technological University, Singapore \\
              \email{mingyuan001@e.ntu.edu.sg}           
          \and
          Zhongang Cai \at
              S-Lab, Nanyang Technological University, Singapore \\
              \email{caiz0023@e.ntu.edu.sg}           
          \and
          Liang Pan \at
              S-Lab, Nanyang Technological University, Singapore \\
              \email{liang.pan@ntu.edu.sg}           
          \and
          Fangzhou Hong \at
              S-Lab, Nanyang Technological University, Singapore \\
              \email{fangzhouhong820@gmail.com}           
          \and
          Xinying Guo \at
              Nanyang Technological University, Singapore \\
              \email{XGUO012@e.ntu.edu.sg}           
          \and
          Lei Yang \at
              SenseTime, China \\
              \email{yanglei@sensetime.com}           
          \and
          Ziwei Liu \at
              S-Lab, Nanyang Technological University, Singapore \\
              \email{ziwei.liu@ntu.edu.sg}           
          \and
          * Equal Contributions
}

\date{Received: date / Accepted: date}

\maketitle

\begin{abstract}

Human motion modeling is important for many modern graphics applications, which typically require professional skills.
In order to remove the skill barriers for laymen, recent motion generation methods can directly generate human motions conditioned on natural languages.
However, it remains challenging to achieve diverse and fine-grained motion generation with various text inputs.
To address this problem, we propose \textbf{\name}, the first diffusion model-based text-driven motion generation framework, which demonstrates several desired properties over existing methods.
\emph{1) Probabilistic Mapping}. Instead of a deterministic language-motion mapping, \name generates motions through a series of denoising steps in which variations are injected. 
\emph{2) Realistic Synthesis}. \name excels at modeling complicated data distribution and generating vivid motion sequences.
\emph{3) Multi-Level Manipulation}. \name responds to fine-grained instructions on body parts, and arbitrary-length motion synthesis with time-varied text prompts. 
Our experiments show \name outperforms existing SoTA methods by convincing margins on text-driven motion generation and action-conditioned motion generation. A qualitative analysis further demonstrates \name's controllability for comprehensive motion generation. Homepage: \url{https://mingyuan-zhang.github.io/projects/MotionDiffuse.html}

\keywords{Motion Synthesis \and Conditional Motion Generation \and Diffusion Model \and Text-driven Generation}
\end{abstract}

\section{Introduction}\label{sec1}

\begin{figure*}[t]
    \centering
    \includegraphics[width=\linewidth]{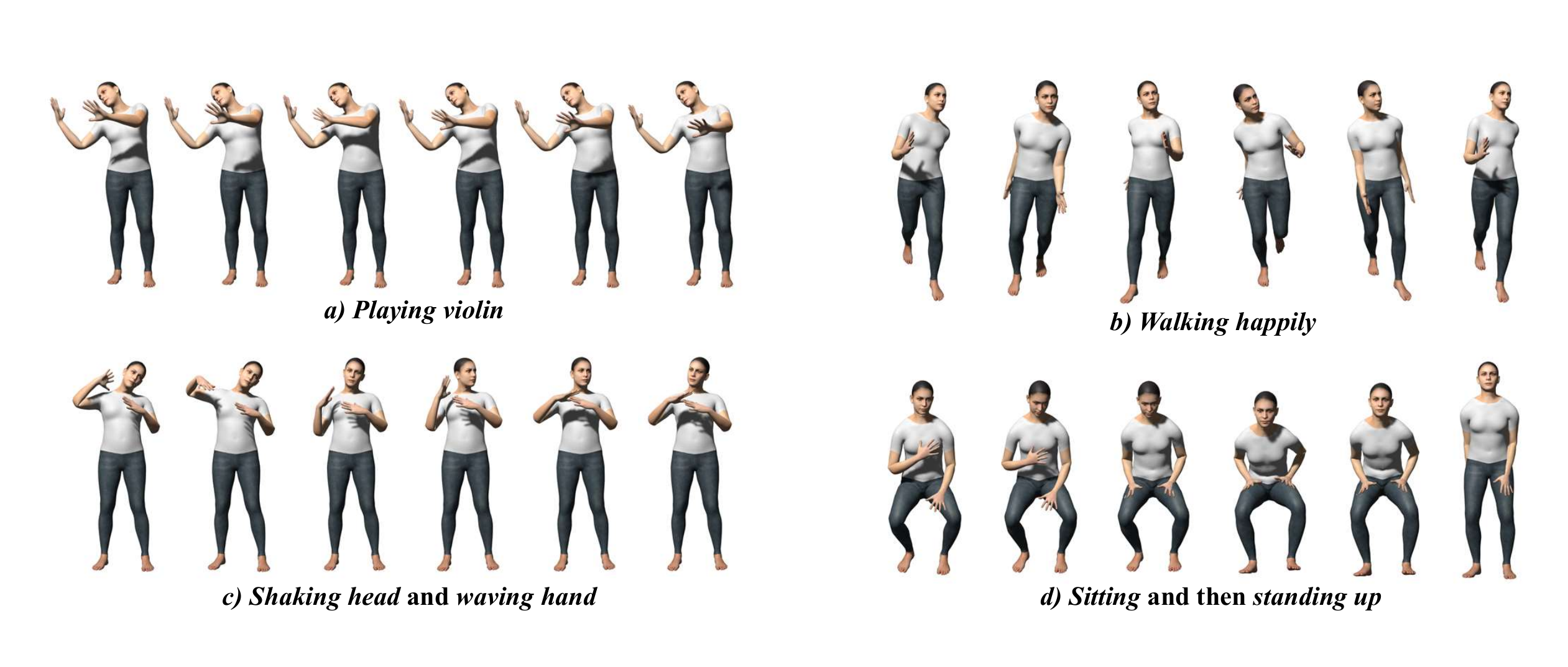}
    \caption{\name is a diffusion model-based text-driven motion generation method that features \emph{1) Probabilistic Mapping 2) Realistic Synthesis} that results in highly diverse motions with high-fidelity shown in a)-d), and \emph{3) Multi-Level Manipulation} that empowers comprehensive motion generation such as that in c) where multiple body parts are involved and d) where time-varied prompts are given.}
    \label{fig:pipeline}
\end{figure*}


Human motion modeling is a critical component of animating virtual characters to imitate vivid and rich human movements, which has been a vital topic for many applications, such as film-making, game development, and virtual YouTuber animation.
To mimic human motions, virtual characters should be capable of moving around naturally, reacting to environmental stimuli, and meanwhile expressing sophisticated emotions.
Despite decades of exciting technological breakthroughs, it requires sophisticated equipment (e.g., expensive motion capture systems) and domain experts to produce lively and authentic body movements.
In order to remove skill prerequisites for layman users and potentially scale to the mass audience, it is vital to create a versatile human motion generation model that could produce diverse, easily manipulable motion sequences.



Various condition signals, including pre-defined motion categories~\citep{guo2020action2motion,petrovich2021action,cervantes2022implicit}, music pieces~\citep{huang2020dance,li2020learning,li2021ai,zhuang2020music2dance,siyao2022bailando}, and natural language ~\citep{lin2018generating,ahuja2019language2pose,ghosh2021synthesis,petrovich2022temos}, have been leveraged in previous human motion generation methods.
Among them, natural language is arguably the most user-friendly and convenient input format for motion sequence synthesis, and hence we focus on text-driven motion generation in this work.
Recently, TEMOS~\citep{petrovich2022temos} utilizes KIT Motion-Language MoCap dataset~\citep{plappert2016kit} to demonstrate fine-grained trajectory synthesis.
However, it does not support stylizing the generated motions and, therefore, could not achieve high diversity.
MotionCLIP~\citep{tevet2022motionclip} could generate stylized motions, but it is still limited to short text inputs and fails to handle complicated motion descriptions.
In addition, they~\citep{petrovich2022temos,tevet2022motionclip} typically only accept a single text prompt, which greatly limits users' creativity.



To tackle the aforementioned challenges, we propose \textbf{\name}, a versatile and controllable motion generation framework that could generate diverse motions with comprehensive texts.
Inspired by the recent progress of the text-conditioned image generation~\citep{dhariwal2021diffusion,nichol2021improved,nichol2021glide,ramesh2022hierarchical}, we propose to incorporate Denoising Diffusion Probabilistic Models (DDPM)~\citep{ho2020denoising} into motion generation.
Unlike classical DDPM which is only capable of fixed-size generation, we propose a Cross-Modality Linear Transformer to achieve motion synthesis with an arbitrary length depending on the motion duration. 
Instead of learning a direct mapping between the text space and the motion space~\citep{tevet2022motionclip}, we propose to guide the generation pipeline with input texts softly, which could significantly increase the diversity of the generation results.
To maintain the uncertainties in the denoising process, we process the noise terms conditioned on the input texts by several transformer decoder layers for each denoising step.
In this way, the text conditions would not dominate the motion generation in a deterministic way, which facilitates generating diverse motion sequences from the driving texts.



Furthermore, \name can achieve body part-independent control with fine-grained texts.
Specifically, to accommodate the human body structures, \name divides the whole-body motion into several near-independent parts(\eg upper body and lower body).
Based on the fine-grained body parts definition, we propose `noise interpolation' to separately control different body parts while taking their correlations into consideration.
Moreover, to synthesize arbitrary-length motion sequences, we propose a new sampling method to denoise several overlapped sequences simultaneously.
Specifically, \name first gets results from each sequence independently and then mixes them with correction terms.
Different from auto-regressive inference schemes that often require many long motion sequences for training, \name is capable of modeling the correlations between continuous actions without introducing additional training costs.


We perform both extensive qualitative experiments on popular benchmarks, and quantitative evaluation on comprehensive motion generation. Firstly, we demonstrate significant improvements on text-driven motion generation over the current art on HumanML3D and KIT-ML. Secondly, to illustrate the superiority of \name, we directly apply it to action-conditioned motion generation tasks, and it outperforms all existing works on the HumanAct12 dataset and the UESTC datasets. Furthermore, we also demonstrate more possibilities of \name by conditioning the model on mixed control signals that allow body-part level manipulation and long motion generation.

In summary, our proposed \name has several desired properties over prior arts: 

\begin{itemize}
    \item \emph{Probabilistic Mapping.} Benefiting from our new formulation of motion generation, where the DDPM is incorporated in, \name can be conditioned on text descriptions to generate motions in a probabilistic style, naturally leading to high diversity.
    \item \emph{Realistic Synthesis.} The careful design of the architecture allows \name to synthesize high-fidelity motion sequences and achieves state-of-the-art on two conditional motion generation tasks. 
    \item \emph{Multi-Level Manipulation.} With the extended design, \name handles fine-grained text descriptions that mobilize the entire body (\eg `a person is drinking water while walking') and time-varied signals (\eg `a person is walking and then running').
\end{itemize}

\section{Related Work}\label{sec2}

\subsection{Motion Generative Model}\label{sec2_1}
Motion Generation has been studied for decades. Some early works focus on unconditional motion generation~\citep{rose1998verbs,ikemoto2009generalizing,mukai2005geostatistical}. While some other works try to predict future motions given an initial pose or a starter motion sequence~\citep{futrelle1978extraction,gavrila1999visual,o1980model}. Statistical models such as PCA~\citep{ormoneit2005representing}, Motion Graph~\citep{min2012motion} are applied for these purposes. 

With the rapid development of Deep Learning (DL) techniques, more generative architectures occur and flourish. Previous works can be broadly divided into four groups: 1) Variational Auto Encoder (VAE); 2) Generative Adversarial Networks (GAN); 3) Normalization Flow Network; 4) Implicit Neural Representations.

VAE~\citep{kingma2013auto} is one of the most commonly used generative models in motion synthesis. \cite{yan2018mt} and \cite{aliakbarian2020stochastic} regard the motion generation task as predicting a small future motion sequence with the given small current motion sequence. They use VAE to encode the pair of current sequence and future sequence and then reconstruct the future one. ACTOR~\citep{petrovich2021action} proposes a transformer-based encoder and decoder architecture. Transformer Encoder Layers and Transformer Decoder Layers~\citep{vaswani2017attention} are the basic blocks to build up a motion encoder and a motion decoder. This architecture is also employed in later works~\citep{tevet2022motionclip,hong2022avatarclip,petrovich2022temos}. 

GAN~\citep{goodfellow2014generative} introduces an auxiliary module, discriminator network, to justify the quality and validity of generated samples.
Some works focus on proposing appropriate discriminator networks for motion generation to improve the synthesis quality~\citep{barsoum2018hp,harvey2020robust,wang2020learning}. HP-GAN~\citep{barsoum2018hp} attempts to supervise the motion prediction results without the specific ground truth. Therefore, a data-driven discriminator is involved in learning a motion prior, which is used to justify the prediction quality. \cite{harvey2020robust} target solving the blurriness of the predicted motion in the Motion In-between task and propose two discriminators for both short-term critic and long-term critic. \cite{wang2020learning} build up a cyclic pipeline. With the help of a discriminator, the proposed pipeline can generate both class-specific and mixed-class motion sequences.

Normalization Flow Network~\citep{dinh2014nice} has a long history and has been studied extensively for image synthesis~\citep{dinh2016density,kingma2018glow}. This kind of architecture builds up a reversible neural network and will map the input data into a multi-dimensional Gaussian distribution. Hence, we can generate an initially random vector from this distribution and feed them into the reversed network to generate motion samples. Inspired by the success of GLOW~\citep{kingma2018glow}, MoGlow~\citep{henter2020moglow} proposes an auto-regressive normalization network to model motion sequences. History features from an LSTM model~\citep{hochreiter1997long} serve as the condition of the flow network, which predicts the next pose.

Recently, another generative model has attracted much attention with the considerable success achieved by NeRF~\citep{mildenhall2020nerf, jain2021putting} in rendering realistic images. Implicit Neural Representations (INR) are a series of neural networks that optimize their parameters to fit one sample instead of the whole distribution. One principal advantage is that this technique has superb generalization ability on spatial or temporal dimensions. For example, \cite{cervantes2022implicit} propose an implicit scheme, which simultaneously models action category and timestamp. Similar to the original NeRF, the timestamp is represented by sinusoidal values. After supervised training, the proposed method can generate a variable-length motion sequence for each action category.   

This paper proposes a new motion generation pipeline based on the Denoising Diffusion Probabilistic Model (DDPM)~\citep{ho2020denoising}. One of the principal advantages of DDPM is that the formation of the original motion sequence can be retained. It means that we can easily apply more constraints during the denoising process. In the later sections, we will explore more potential of DDPM in different types of conditions. Besides, benefiting from this nature, DDPM can generate more diverse samples.

\subsection{Conditional Motion Generation}\label{sec2_2}
The increasing maturity of various generative models stimulates researchers' enthusiasm to study conditional motion generation. For example, some works~\citep{guo2020action2motion,petrovich2021action,cervantes2022implicit} aim at synthesizing motion sequences of several specific categories. Action2Motion~\citep{guo2020action2motion} builds up a recurrent conditional VAE for motion generation. Given history memory, this model predicts the next pose under the constraints of the action category. ACTOR~\citep{petrovich2021action} also uses VAE for random sampling. Unlike Action2Motion, ACTOR embeds the whole motion sequence into the latent space. This design avoids the accumulative error in the recurrent scheme. Besides, ACTOR proposes a Transformer-based motion encoder and decoder architecture. This structure significantly outperforms recurrent methods. \cite{cervantes2022implicit} attempt to model motion sequence with implicit functions, which can generate motion sequences with varied lengths.

Another significant conditional motion generation task is music to dance. This task requires that the generated motion has beat-wise connectivity, is a specific kind of dance, or can express similar content with the music. Many works attempt to embed the music feature and motion feature into a joint space~\citep{lee2019dancing, sun2020deepdance, li2020learning, li2021ai}. Unlike direct feature embedding, Bailando~\citep{siyao2022bailando} proposes a two-stage dance generator. It first learns a quantized codebook of meaningful dance pieces and then attempts to generate the whole sequence with a series of elements from a codebook.

Similar to music-to-dance, text-driven motion generation can be regarded as learning a joint embedding of text feature space and motion feature space. There are two major differences. The first one is that language commands correlate more with the human body. Therefore, we expect to control each body part accurately. The second difference is that text-driven motion generation contains a vast range of motions. Some descriptions are direct commands to a specific body part, such as ``touch head''. Some describes arbitrary concepts like ``playing the violin''. Such huge complexity of motions brings many difficulties to the architecture design. Recently, many works have proposed text-driven motion generation pipelines. Most of them are deterministic generation~\citep{ahuja2019language2pose,ghosh2021synthesis,tevet2022motionclip}, which means they can only generate a single result from the given text. TEMOS ~\citep{petrovich2022temos} introduces the VAE architecture into this task. It can generate different motion sequences given one text description. However, these methods attempt to acquire a joint embedding space of motion and natural language. This design significantly compresses the information from text. Therefore, these works can hardly generate correct motion sequences from a detailed description. \cite{guo2022generating} proposes an auto-regressive pipeline. It first encodes language descriptions into features and then auto-regressively generates motion frames conditioned on the text features. However, this method is hard to capture the global relation due to the auto-regressive scheme. Moreover, the generation quality is inferior. Instead, our proposed \name softly fuses text features into generation and can yield the whole sequence simultaneously. The experiment results prove the superiority of our design.

\subsection{Motion Datasets}\label{sec2_3}
Human motion modeling has been a long-standing problem in computer vision and computer graphics. With the advent of deep learning, data has become increasingly important for training neural networks that perceive, understand, and generate human motions. 

A common form of datasets containing videos of human subjects are recorded with annotations such as 2D keypoints \citep{jhuang2013towards, andriluka2018posetrack}, 3D keypoints \citep{ionescu2013human3, joo2015panoptic, mehta2017monocular, trumble2017total, li2021ai} and statistical model parameters \citep{Yu2020HUMBIAL, patel2021agora, cao2020long, cai2021playing, cai2022humman}. Action labels are also a popular attribute of datasets for human action understanding that contains human-centric actions~\citep{kuehne2011hmdb, soomro2012ucf101, karpathy2014large, gu2018ava, shao2020finegym, chung2021haa500}, interaction~\citep{carreira2019short, monfort2019moments, zhao2019hacs}, fine-grained action understanding~\citep{gu2018ava, shao2020finegym, chung2021haa500} and 3D data~\citep{liu2019ntu}. For the action-conditioned motion generation task, HumanAct12~\citep{guo2020action2motion}, UESTC~\citep{ji2018large}, and NTU RGB+D~\citep{liu2019ntu} are three commonly used benchmarks. However, the above-mentioned datasets do not provide paired sophisticated semantic labels to the motion sequences.

KIT~\citep{plappert2016kit} contains motion capture data annotated with detailed descriptions. \cite{zhang2020kinematic} recruit actors and actresses to record body movements expressing emotions. Recently, BABEL~\citep{punnakkal2021babel} and HumanML3D~\citep{guo2022generating} re-annotates AMASS~\citep{mahmood2019amass}, a large scale motion capture dataset, with English language labels.

In this paper, we use the HumanML3D dataset and KIT dataset to evaluate the proposed methods for the text-driven motion generation task. HumanAct12 and UESTC are used to demonstrate the wide applicability of the proposed pipeline. Furthermore, we use the BABEL dataset for additional applications.

\section{Methodology}\label{sec3}

We present a diffusion model-based framework, \textbf{\name}, for high-fidelity and controllable text-driven motion generation. We first give the problem definition, settings of the original text-driven motion generation in Section \ref{sec3_1}. After that, we provide an overall illustration of the proposed \name in Section \ref{sec3_2}, followed by introducing the diffusion model in Section \ref{sec3_3} and the transformer-based architecture in Section \ref{sec3_4}. Finally, the inference strategy is illustrated for the fine-grained generation scenarios in Section \ref{sec3_5}.

\subsection{Preliminaries}\label{sec3_1}

The motion sequence $\Theta$ is an array of $(\theta_i)$, $i \in \{1,2,\dots,F\}$, where $\theta_i \in \mathbb{R}^{D}$ represents the pose state in the $i$-th frame, and $F$ is the number of frames. The representation of each pose state $\theta_i$ is distinct in different datasets. It generally contains joint rotation, joint position, joint velocity, and foot contact conditions. Our proposed \name is robust to the various motion representations. Therefore, we do not specify the components of $\theta_i$ in this section, and leave the details in Section \ref{sec4}. 

For standard Text-driven Motion Generation, the training datasets consist of $(\theta_i, \textrm{text}_i)$ pairs, where $\textrm{text}_i$ is the language description of motion sequence $\theta_i$. During inference, given a set of descriptions $\{\textrm{text}_i\}$, we are requested to generate motion sequences conditioned on the given descriptions. This task can also be regarded as a text-to-motion translation (T2M) task. We will use this abbreviation below.

An related task is Action-conditioned Motion Generation. Given a pre-defined action category set, models are supposed to fit the data distribution and synthesize motion sequences of each category. Annotated data in this task can be represented as ${(y_i, \Theta_i)}$, where $y_i$ is the category index of $i$-th data, $\Theta_i$ is the motion sequence of $i$-th data. In this paper, we replace $y_i$ by its semantic description $\textrm{text}_i$. Then we can use the same pipeline as in the T2M task.

\subsection{Pipeline Overview}\label{sec3_2}

\begin{figure*}[t]
    \centering
    \includegraphics[width=\linewidth]{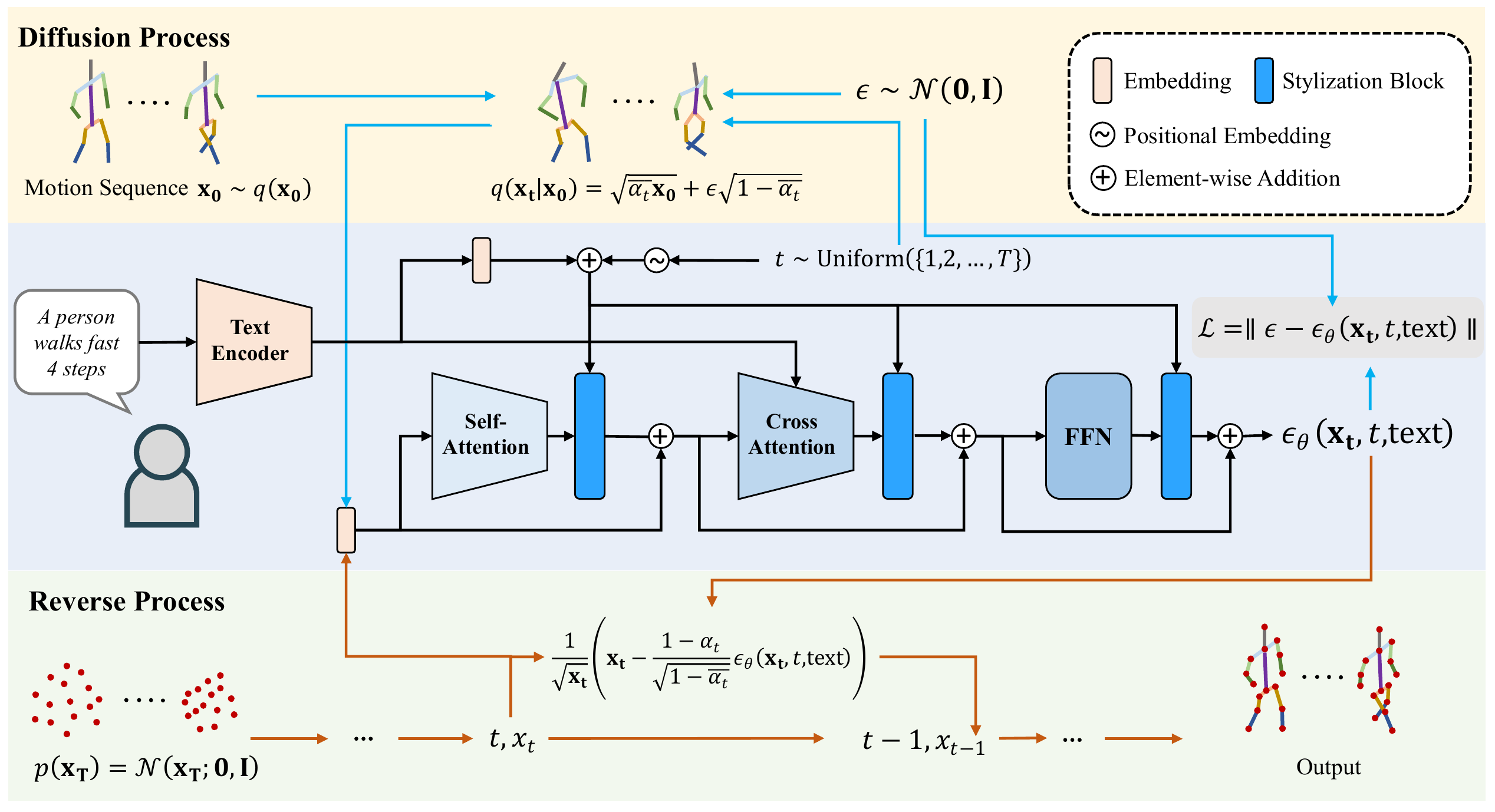}
    \caption{\textbf{Overall Pipeline of the proposed \name}. The colors of the arrows indicate different stages: blue for training, red for inference, and black for both training and inference.}
    \label{fig:pipeline}
\end{figure*}

Following the literature on the diffusion model in the image synthesis field~\citep{ho2020denoising}, we first build up a text-conditioned motion generation pipeline using a denoising diffusion probabilistic model (DDPM). This model is the basis of our proposed \name. For the denoising process, we propose a Cross-Modality Linear Transformer to process input sequences conditioned on the given text prompts. Beyond the direct application of text-driven motion generation, we take one step further to explore methods that are conditioned on motion representation during denoising. Specifically, we experiment with two types of additional signals: part-aware text controlling and time-varied controlling. The former assigns different text conditions to different body parts so that we can accurately control each part of the body and generate more complicated motion sequences. The latter divides the whole sequence into several parts and assigns independent text conditions for each interval. Therefore, we can synthesize arbitrary-length motion sequences that incorporate several actions. These two kinds of conditions significantly expand the capability of \name. The overall pipeline is shown in Figure~\ref{fig:pipeline}. We introduce each part of this architecture in the following subsections.

\subsection{Diffusion Model for Motion Generation}\label{sec3_3}

Generative Adversarial Networks (GANs) involve a discriminator to improve the generation quality in an adversarial manner. GANs are typically challenging to train, especially for conditional motion generation tasks. Implicit Functions use Multi-Layer Perceptron (MLP) to fit motion sequences. This neat architecture is easily trained on a small number of data but tends to be less generalizable when it is subjected to complicated conditions. Auto-Encoder (AE) and Variational Auto-Encoder (VAE) are the most widely used approaches in text-driven motion generation~\citep{ghosh2021synthesis,petrovich2022temos}. Previous works learn a joint embedding of motion sequences and languages that explicitly apply the text condition in the deterministic language-motion mapping. However, high-level text features typically contain insufficient fine-grained details to guide the generation of subtly different motion. Hence, directly linking text embedding to motion embedding results in the limited diversity of the generated motions.

To tackle the problem, we build our text-driven motion generation pipeline based on diffusion models. Diffusion Models~\citep{ho2020denoising,dhariwal2021diffusion,nichol2021improved,nichol2021glide} are a new class of generative models.  
A probabilistic model is learned to gradually denoises a Gaussian noise to generate a target output, such as a 2D image or 3D point cloud.
Formally, diffusion models are formulated as $p_{\theta}(\mathbf{x}_0)\,:=\,\int{p_{\theta}(\mathbf{x}_{0:T})\,d{\mathbf{x}_{1:T}}}$, where $\mathbf{x}_0 \sim q(\mathbf{x}_0)$ is the real data, and $\mathbf{x}_{1},\cdots,\mathbf{x}_{T}$ are the latent data.
They generally have a diffusion process and a reverse process.
To approximate posterior $q(\mathbf{x}_{1:T} \vert \mathbf{x}_0)$, the diffusion process follows a Markov chain to gradually add Gaussian noise to the data until its distribution is close to the latent distribution $\mathcal{N}(\mathbf{0}, \mathbf{I})$, according to variance schedules given by $\beta_t$:
\begin{equation}
    \begin{aligned}
        &q(\mathbf{x}_{1:T} \vert \mathbf{x}_0) \,:=\, \prod_{t=1}^{T} q(\mathbf{x}_t \vert \mathbf{x}_{t-1}), \\
        &q(\mathbf{x}_t \vert \mathbf{x}_{t-1}) \,:=\, \mathcal{N}(\mathbf{x}_t; \sqrt{1-\beta_t}\mathbf{x}_{t-1}, \beta_t\mathbf{I}).
    \end{aligned}
\end{equation}

The reverse process $p_{\theta}(\mathbf{x}_{0:T})$ is also a Markov chain that predicts and eliminates the noise with learned Gaussian transitions starting at $p(\mathbf{x}_{T}) = \mathcal{N}(\mathbf{x}_{T};\mathbf{0}, \mathbf{I})$:
\begin{equation}
\label{eq:ddpm_reverse}
    \begin{aligned}
        &p_{\theta}(\mathbf{x}_{0:T}) \,:=\, p(\mathbf{x}_{T}) \prod_{t=1}^{T}p_{\theta}(\mathbf{x}_{t-1} \vert \mathbf{x}_{t}),\\
        &p_{\theta}(\mathbf{x}_{t-1} \vert \mathbf{x}_{t}) \,:=\, \mathcal{N}(\mathbf{x}_{t-1}; \mu_\theta(\mathbf{x}_{t}, t), \Sigma_\theta(\mathbf{x}_{t}, t)).
    \end{aligned}
\end{equation}

To accomplish the reverse process of the diffusion model, we need to construct and optimize a neural network. During training, first we uniformly sample steps $t$ for each ground truth motion $\mathbf{x}_0$ and then generate a sample from 
$q(\mathbf{x}_t \vert \mathbf{x}_0)$. Instead of repeatedly adding noises on $\mathbf{x}_0$, \cite{ho2020denoising} formulate the diffusion process as
\begin{equation}
    q(\mathbf{x}_t \vert \mathbf{x}_0)= \sqrt{\bar{\alpha_t}}\mathbf{x}_0 + \epsilon \sqrt{1-\bar{\alpha_t}}, \epsilon \sim \mathcal{N}(\mathbf{0}, \mathbf{I}),
\end{equation}
where $\alpha_t=1-\beta_t$, $\bar{\alpha_t}=\prod_{s=0}^t \alpha_s$. Hence, we can simply sample a noise $\epsilon$ and then directly generate $\mathbf{x}_t$ by this formulation. Instead of predicting $\mathbf{x}_{t-1}$, here we follow GLIDE~\citep{nichol2021glide} and predict the noise term $\epsilon$. It means that we need to construct a network to fit $\epsilon_{\theta}(\mathbf{x}_t,t,\textrm{text})$. We optimize the model parameters to decrease a mean squared error as
\begin{equation}
    \mathcal{L}=\mathrm{E}_{t \in [1,T], \mathbf{x}_0 \sim q(\mathbf{x}_0),\epsilon \sim \mathcal{N}(\mathbf{0}, \mathbf{I})} [\parallel \epsilon - \epsilon_{\theta}(\mathbf{x}_t,t,\textrm{text}) \parallel].
\end{equation}
This is the only loss we used in model training. To generate samples from the given text description, we denoise the sequence from $p(\mathbf{x}_{T}) = \mathcal{N}(\mathbf{x}_{T};\mathbf{0}, \mathbf{I})$. Equation ~\ref{eq:ddpm_reverse} shows that we need to estimate $\mu_\theta(\mathbf{x}_{t}, t, \textrm{text})$ and $\Sigma_\theta(\mathbf{x}_{t}, t, \textrm{text})$. To simplify the problem, we set $\Sigma_\theta(\mathbf{x}_{t}, t, \textrm{text})$ as a constant number $\beta_t$. $\mu_\theta(\mathbf{x}_{t}, t, \textrm{text})$ can be estimated as
\begin{equation}
    \mu_\theta(\mathbf{x}_{t}, t, \textrm{text}) = \frac{1}{\sqrt{\mathbf{x}_t}}(\mathbf{x}_t-\frac{1-\alpha_t}{\sqrt{1-\bar{\alpha_t}}} \epsilon_{\theta}(\mathbf{x}_t,t,\textrm{text})).
\end{equation}
Therefore we can denoise the motion sequence step by step and finally get a clean motion sequence, which is conditioned on the given text.

\subsection{Cross-Modality Linear Transformer}\label{sec3_4}

\begin{figure}[t]
    \centering
    \includegraphics[width=\linewidth]{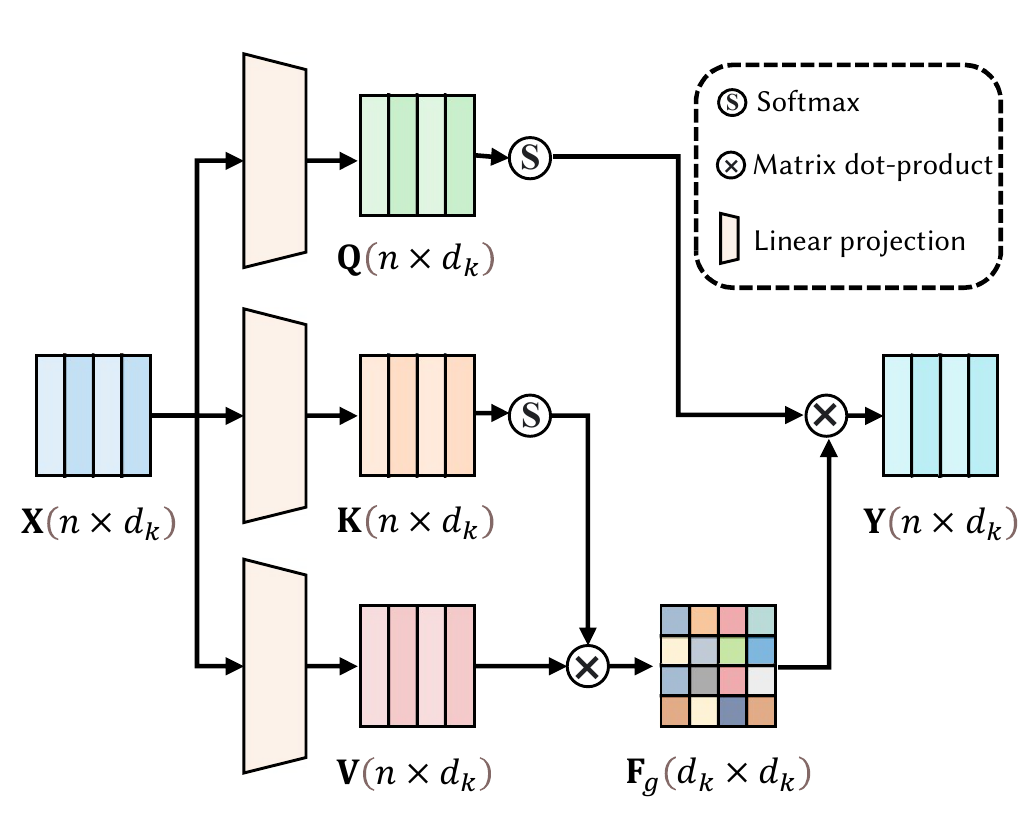}
    \caption{\textbf{Architecture} of our Linear Self-attention.}
    \label{fig:arch}
\end{figure}

In Section~\ref{sec3_3}, we illustrate diffusion models as motion generators and a neural network $\epsilon_{\theta}(\mathbf{x}_t,t,\textrm{text})$ is essential for denoising steps. In this section, we will introduce the design of $\epsilon_{\theta}(\mathbf{x}_t,t,\textrm{text})$ in our proposed \name.

Previous works~\citep{ho2020denoising,dhariwal2021diffusion,nichol2021improved,nichol2021glide} mainly utilize UNet-like structure as the denoising model. However, the target motion sequences are variable-length in the motion generation task, making convolution-based networks unsuitable. Therefore, we propose a Cross-Modality Linear Transformer, as shown in Figure ~\ref{fig:pipeline}. Similar to the machine translation task, our proposed model includes a text encoder and a motion decoder. To meet the requirement of the diffusion models, we further customize each layer of the motion decoder.

\paragraph{Text Encoder} Here we directly use classical transformer~\cite{vaswani2017attention} to extract text features. Specifically, the input data first passes through an embedding layer to get the embedding feature from raw text and then is further processed by a series of transformer blocks. Each block contains two components: a multi-head attention module (MHA) and a feed-forward network (FFN). Suppose the input feature is $\mathbf{X} \in \mathbb{R}^{n \times d}$ ($n$ denotes the number of elements and $d$ denotes the element feature dimension), MHA extracts query feature vectors $\mathbf{Q} \in \mathbb{R}^{n \times d}$, key feature vectors $\mathbf{K} \in \mathbb{R}^{n \times d}$, and value feature vectors $\mathbf{V} \in \mathbb{R}^{n \times d}$ as:

\begin{equation}
\label{eq:linear}
    \textbf{Q} = W_q\, \textbf{X},\;\;\;
    \textbf{K} = W_k\, \textbf{X},\;\;\;
    \textbf{V} = W_v\, \textbf{X},
\end{equation}
where $W_q$, $W_k$ and $W_v$ are the corresponding linear projections to generate $\mathbf{Q}$, $\mathbf{K}$ and $\mathbf{V}$, respectively. 
%
The value features are then aggregated with attention weights $\mathbf{A}  \in \mathbb{R}^{n \times n}$:
\begin{equation}
\label{eq:self_att}
    \mathbf{A}=\operatorname{softmax}(\frac{\mathbf{Q}\otimes \mathbf{K}^{\top}}{\sqrt{d}} ),\;\;\; \mathbf{Y}=\mathbf{A} \otimes \mathbf{V},
\end{equation}
where $\mathbf{Y}\in \mathbb{R}^{n \times d}$ is the output of MHA modules, $d$ is the dimension of each element in $\mathbf{X}$, and $\otimes$ denotes the matrix multiplication. The multi-head mechanism divides the input vector into several parts, which pass through the process in Equation~\ref{eq:linear} and~\ref{eq:self_att} independently. The outputs are concatenated so that the dimension remains unchanged. A residual connection is applied between input and output of the MHA modules. This feature is further processed by FFN, which contains three linear transformations and two GELU~\citep{hendrycks2016gaussian} layers between them.

To enhance the generalization ability, we use parameter weights in  CLIP~\citep{radford2021learning} to initialize the first several layers. This part of the parameters is frozen and will not be optimized in the later training.

\paragraph{Linear Self-attention.} This module aims at enhancing motion features by modeling correlations between different frames. The principal advantage of self-attention is to get an overview of the input sequence and is thus beneficial to estimating the injected noise $\epsilon$. However, the time complexity of calculating attention weight $\mathbf{A} \in \mathbb{R}^{n \times n}$ is $\mathcal{O}(n^2d)$. When the target motion sequence length increases, the time cost increases quadratically. In the T2M task, the length can be several hundred, which leads to low speed. Hence, we adopt Efficient Attention~\citep{shen2021efficient} to speed up the self-attention module. Instead of calculating pair-wise attention weights, efficient attention generates global feature map $\mathbf{F}_g \in \mathbb{R}^{d_k \times d_k}$, where $d_k$ is the dimension of feature after multi-head split:
\begin{equation}
    \mathbf{F}_g = \operatorname{softmax}(\mathbf{K}^{\top}) \otimes \mathbf{V},\;\;\; \mathbf{Y}=\operatorname{softmax}(\mathbf{Q}) \otimes \mathbf{F}_g.
\end{equation}
The time complexity of these two steps is $\mathcal{O}(d_k^2nk)=\mathcal{O}(dd_kn)$, where $n$ is the number of the elements in the sequences, $k$ is the number of heads of self-attention.


Another advantage of efficient attention for the diffusion model is that feature map $\mathbf{F}_g$ explicitly aggregates global information while classical self-attention focus more on pair-wise relation. Global information gives more clues about the semantic meaning of the motion sequence than pair-wise one. The experiment results also prove this conclusion.

\paragraph{Linear Cross-attention.} Cross-attention replaces $\mathbf{X}$ in $\mathbf{K}$ and $\mathbf{V}$ calculation by the text feature. Other formulations are the same as Linear Self-attention. Text features are injected into motion sequences in this process to generate motion conditioned on the given text. 

\paragraph{Stylization Block.} In each denoising step, the output is conditioned on the given text and the timestamp $t$. Linear Cross-attention fuses the text features into motion sequences. We need another Stylization Block component to bring timestamp $t$ to the generation process. This block is applied after each Linear Self-attention block, Linear Cross-attention block, and FFN block.

Similar to GLIDE~\citep{nichol2021glide}, we first get a text embedding $e_{\textrm{text}}$ by a linear transformation on the text features and a timestamp embedding $\mathbf{e_t}$ by positional embedding~\citep{vaswani2017attention}. These two terms are summed together into one single vector $\mathbf{e}$. Given the original output $\mathbf{Y}$ from other blocks, the Stylization block will process the feature as:
\begin{equation}
\mathbf{B} = \psi_b(\phi(\mathbf{e})),\;\;\; \mathbf{W} = \psi_w(\phi(\mathbf{e})),\;\;\; \mathbf{Y}^{\prime}=\mathbf{Y} \cdot \mathbf{W} + \mathbf{B},
\end{equation}
where $(\cdot)$ denotes Hadamard product, $\mathbf{Y}^{\prime}$ is the output of stylization blocks. $\psi_b, \psi_w, \phi$ denote three different linear projections. In classical transformers, the output from each block is added to the original input as a residual connection, as shown in Figure ~\ref{fig:pipeline}. In \name, these outputs pass through stylization blocks and are added to the information. This modification enables our proposed method to know the timestamp $t$. 

\subsection{Fine-grained Controlling}\label{sec3_5}

To enrich the capability of \name, we explore the properties of both the motion representation and the denoising process of DDPM. Unlike VAE, the generated motion sequence is in its explicit form instead of being compressed in the latent space. This characteristic of DDPM-based motion generation allows more operations to be applied to this motion sequence to increase the manipulability.

\paragraph{Body Part-independent Controlling.}
Due to the lack of diversity in text descriptions, we cannot achieve accurate motion control for each body part from text descriptions only. For example, the prompt `a person is running and waving left hand' is highly challenging to the model because the expected motion sequence is significantly far from the training distribution. Even if we manually split the original description into two independent ones: `a person is running' for lower limbs, and `a person is waving left hand' for upper limbs, it is still difficult for the model to generate correct motions. An intuitive solution for this situation is to separately generate two motion sequences and combine the upper-limb motion of the first sequence and the lower-limb motion of the second sequence. This simple solution mitigates the problem to some extent. However, it ignores the correlation between these two parts. Specifically for `running and waving left hand', the frequencies of the two motions should match. Otherwise, the motion generated by this naive method appears unnatural. To better solve this problem, we propose a body part-independent controlling scheme.

Recall that, during the denoising process, our diffusion model predicts the noise term $\epsilon_{\theta}(\mathbf{x}_t,t,\textrm{text}) \in \mathbb{R}^{F \times D}$, where $F$ represents the number of frames, $D$ denotes the dimension of each pose state, which includes translation and rotations of body joints. This noise term determines the denoising direction of the whole body. 

Inspired by the application of the latent code interpolation, here we propose `noise interpolation' to separately control the different parts of the human body. Suppose we have $n$ text descriptions $\{\textrm{text}_i\}$ for different body parts $\{s_i\}$. We want to calculate the noise term $\epsilon=\{\epsilon^{\textrm{joint}}_i\}, i \in [1, m]$, where $\epsilon^{\textrm{joint}}_i$ represents the noise term for the $i$-th body part, $m$ denotes the number of partition. We first estimate the noise $\epsilon^{\textrm{part}}_i=\epsilon_{\theta}(\mathbf{x}_t,t,\textrm{text}_i), \epsilon^{\textrm{part}}_i \in \mathbb{R}^{F \times D}$. An intuitive method for combining these terms is $\epsilon^{\textrm{part}}=\sum_{i=1}^m \epsilon^{\textrm{part}}_i \cdot M_{i}$, where $M_i \in \{0,1\}^D$ is a binary vector to show which body part should we focus. $(\cdot)$ denotes the Hadamard product, and here we ignore the broadcast in computation for simplicity. Although this method succeeds to some extent, the direct ignoring of some parts in $\epsilon^{\textrm{part}}_i$ will increase the combination difficulty and lead to low-quality generation results. Therefore, we add a correction item for smoothing interpolation:
\begin{equation}
\overline{\epsilon}^{\textrm{part}}=\sum_{i=1}^m \epsilon^{\textrm{part}}_i \cdot M_{i} + \lambda_1 \cdot \nabla ( \sum\limits_{1 \leq i,j \leq m} \Vert \epsilon^{\textrm{part}}_i - \epsilon^{\textrm{part}}_j  \Vert),
\end{equation}
where $\nabla$ denotes gradient calculation, $\lambda_1$ is a hyper-parameter to balance these two items. This correction item enforces a smoother denoising direction so that the motion of different body parts will be more natural.

\paragraph{Time-varied Controlling} Long-term motion generation plays a vital role in real-world applications. Previous works mainly focus on motion generation with a single type of motion. Auto-regressive methods~\citep{henter2020moglow,guo2022generating} have solved this problem with satisfactory performance. However, none of them are capable of synthesizing different actions in a continuous manner. Benefiting from the nature of DDPM, here we propose another sampling method to meet this requirement.

Recall that we are given an array $\{\textrm{text}_{i,j}, [l_{i,j}, r_{i,j}]\}$, $i \in [1, m]$, where $m$ is the number of intervals. Similar to the method we proposed in the previous paragraph, we first estimate the noise term $\epsilon^{\textrm{time}}_i$ for $i$-th interval independently. Suppose the overall length of the target sequence is $F^{\prime}$. By padding zeros, we extend each noise term into the same dimension $F^{\prime} \times D$. Then we interpolate these noises with a correcting term:
\begin{equation}
\overline{\epsilon}^{\textrm{time}}=\sum_{i=1}^m \overline{\epsilon}^{\textrm{time}}_i + \lambda_2 \cdot \nabla ( \sum\limits_{1 \leq i,j \leq m} \Vert \overline{\epsilon}^{\textrm{time}}_i - \overline{\epsilon}^{\textrm{time}}_j  \Vert),
\end{equation}
where $\overline{\epsilon}^{\textrm{time}}_j$ is the padded term from $\epsilon^{\textrm{time}}_i$, $\lambda_2$ is a hyper-parameter.

\section{Experiments}\label{sec4}


We evaluate \name with three categories of experiments: 
text-driven motion generation (Section~\ref{sec4_1}), action-conditioned motion generation (Section~\ref{sec4_2}), 
and motion manipulation (Section~\ref{sec4_3}).
In all the evaluated benchmarks, \name could significantly outperform previous SoTA methods.

\subsection{Text-driven Motion Generation}\label{sec4_1}

\begin{table*}[ht]
\centering
\caption{\textbf{Quantitative results on the HumanML3D test set.} All methods use the real motion length from the ground truth. `$\rightarrow$' means results are better if the metric is closer to the real motions. We run all the evaluation 20 times and $\pm$ indicates the 95\% confidence interval. The best results are in \textbf{bold}.}
\label{tab:humanml3d}
\setlength{\tabcolsep}{1.4mm}
{
\begin{tabular}{lccccccc}
\hline

\multirow{2}{2cm}{\centering Methods} & \multicolumn{3}{c}{\centering R Precision$\uparrow$} & \multirow{2}{1.5cm}{\centering FID$\downarrow$} & \multirow{2}{2.5cm}{\centering MultiModal Dist$\downarrow$} & \multirow{2}{2cm}{\centering Diversity$\rightarrow$} & \multirow{2}{2cm}{\centering MultiModality} \\
& Top 1 & Top 2 & Top 3 \\
\hline
Real motions & $0.511^{\pm .003}$ & $0.703^{\pm.003}$ & $0.797^{\pm.002}$ & $0.002^{\pm.000}$ & $2.974^{\pm.008}$ & $9.503^{\pm.065}$ & -\\ 
\hline

Language2Pose & $0.246^{\pm.002}$ & $0.387^{\pm.002}$ & $0.486^{\pm.002}$ & $11.02^{\pm.046}$ & $5.296^{\pm.008}$ & $7.676^{\pm.058}$ & - \\

Text2Gesture & $0.165^{\pm.001}$ & $0.267^{\pm.002}$ & $0.345^{\pm.002}$ & $7.664^{\pm.030}$ & $6.030^{\pm.008}$ & $6.409^{\pm.071}$ & - \\

MoCoGAN & $0.037^{\pm.000}$ & $0.072^{\pm.001}$ & $0.106^{\pm.001}$ & $94.41^{\pm.021}$ & $9.643^{\pm.006}$ & $0.462^{\pm.008}$ &
$0.019^{\pm.000}$ \\

Dance2Music & $0.033^{\pm.000}$ & $0.065^{\pm.001}$ & $0.097^{\pm.001}$ & $66.98^{\pm.016}$ & $8.116^{\pm.006}$ & $0.725^{\pm.011}$ & $0.043^{\pm.001}$ \\

Guo et al. & $0.457^{\pm.002}$ & $0.639^{\pm.003}$ & $0.740^{\pm.003}$ & $1.067^{\pm.002}$ & $3.340^{\pm.008}$ & $9.188^{\pm.002}$ & $2.090^{\pm.083}$ \\
\hline
Ours & $\mathbf{0.491^{\pm.001}}$ & $\mathbf{0.681^{\pm.001}}$ & $\mathbf{0.782^{\pm.001}}$ & $\mathbf{0.630^{\pm.001}}$ & $\mathbf{3.113^{\pm.001}}$ & $\mathbf{9.410^{\pm.049}}$ & $1.553^{\pm.042}$ \\
\hline
\end{tabular}}
\end{table*}

\begin{table*}[ht]
\centering
\caption{\textbf{Quantitative results on the KIT-ML test set.} All methods use the real motion length from the ground truth.}
\label{tab:kit}
\setlength{\tabcolsep}{1.4mm}
{
\begin{tabular}{lccccccc}
\hline

\multirow{2}{2cm}{\centering Methods} & \multicolumn{3}{c}{\centering R Precision$\uparrow$} & \multirow{2}{1.5cm}{\centering FID$\downarrow$} & \multirow{2}{2.5cm}{\centering MultiModal Dist$\downarrow$} & \multirow{2}{2cm}{\centering Diversity$\rightarrow$} & \multirow{2}{2cm}{\centering MultiModality} \\
& Top 1 & Top 2 & Top 3 \\
\hline
Real motions & $0.424^{\pm .005}$ & $0.649^{\pm.006}$ & $0.779^{\pm.006}$ & $0.031^{\pm.004}$ & $2.788^{\pm.012}$ & $11.08^{\pm.097}$ & -\\ 
\hline

Language2Pose & $0.221^{\pm.005}$ & $0.373^{\pm.004}$ & $0.483^{\pm.005}$ & $6.545^{\pm.072}$ & $5.147^{\pm.030}$ & $9.073^{\pm.100}$ & - \\

Text2Gesture & $0.156^{\pm.004}$ & $0.255^{\pm.004}$ & $0.338^{\pm.005}$ & $12.12^{\pm.183}$ & $6.964^{\pm.029}$ & $9.334^{\pm.079}$ & - \\

MoCoGAN & $0.022^{\pm.002}$ & $0.042^{\pm.003}$ & $0.063^{\pm.003}$ & $82.69^{\pm.242}$ & $10.47^{\pm.012}$ & $3.091^{\pm.043}$ &
$0.250^{\pm.009}$ \\

Dance2Music & $0.031^{\pm.002}$ & $0.058^{\pm.002}$ & $0.086^{\pm.003}$ & $115.4^{\pm.240}$ & $10.40^{\pm.016}$ & $0.241^{\pm.004}$ & $0.062^{\pm.002}$ \\

Guo et al. & $0.370^{\pm.005}$ & $0.569^{\pm.007}$ & $0.693^{\pm.007}$ & $2.770^{\pm.109}$ & $3.401^{\pm.008}$ & $10.91^{\pm.119}$ & $1.482^{\pm.065}$ \\
\hline
Ours & $\mathbf{0.417^{\pm.004}}$ & $\mathbf{0.621^{\pm.004}}$ & $\mathbf{0.739^{\pm.004}}$ & $\mathbf{1.954^{\pm.062}}$ & $\mathbf{2.958^{\pm.005}}$ & $\mathbf{11.10^{\pm.143}}$ & $0.730^{\pm.013}$\\
\hline
\end{tabular}}
\end{table*}

\paragraph{Datasets.}\label{sec4_1_1} KIT Motion Language datset~\citep{plappert2016kit} provides 3911 motion sequences and 6353 sequence-level natural language descriptions. HumanML3D~\citep{guo2022generating} re-annotates the AMASS dataset~\citep{mahmood2019amass} and the HumanAct12 dataset~\citep{guo2020action2motion}. It provides 44970 annotations on 14616 motion sequences. KIT and HumanML3D are two important benchmarks for text-driven motion generation tasks. Following ~\cite{guo2022generating}, we utilize the pretrained text-motion contrastive model.

\paragraph{Evaluation Metrics.}\label{sec4_1_2}
We evaluate all methods with five different metrics. 
1) \textit{Frechet Inception Distance (FID).} Features are extracted from both the generated results and ground truth motion sequences by the pretrained motion encoder. FID is calculated between these two distributions to measure the similarity. 
2) \textit{R Precision.} For each pair of generated sequence and description text, 31 other prompts are randomly selected from the test set. The pretrained contrastive model calculates the average top-k accuracy. This section reports the top-1, top-2, and top-3 accuracies.
3) \textit{Diversity.} The generated sequences from all test texts are randomly split into pairs. Then the average joint differences are calculated in each pair, which serves as the diversity metric.
4) \textit{Multimodality.} As for a single text description, we randomly generate 32 motion sequences. Multimodality measures the differences in joint positions between these homogeneous motion sequences.
5) \textit{Multimodal Distance.} Assisted by the pretrained contrastive model, we can calculate the difference between the text feature from the given description and the motion feature from the generated results, called multimodal distance.

In this section, R Precision and FID are the principal metrics from which we make the most conclusions. Besides, for a fair comparison, we run each evaluation 20 times and report the statistic interval with 95\% confidence.

\paragraph{Implementation Details.}\label{sec4_1_3}
For both HumanML3D and KIT-ML datasets, we build up an 8-layer transformer as the motion decoder. As for the text encoder, we first directly use the text encoder in the CLIP ViT-B/32~\citep{radford2021learning}, and then add four more transformer encoder layers. The latent dimension of the text encoder and the motion decoder are 256 and 512, respectively. As for the diffusion model, the number of diffusion steps is 1000, and the variances $\beta_t$ are linearly from 0.0001 to 0.02. We opt for Adam as the optimizer to train the model with a 0.0002 learning rate. We use 8 Tesla V100 for the training, and there are 128 samples on each GPU, so the total batch size is 1024. The total number of iterations is 40K for KIT-ML and 100K for HumanML3D.

Following \cite{guo2022generating}, pose states in this series of experiments mainly contain seven different parts: $(r^{va},r^{vx}, r^{vz},r^h, \mathbf{j}^p, \mathbf{j}^v, \mathbf{j}^r)$. Here Y-axis is perpendicular to the ground. $r^{va},r^{vx}, r^{vz} \in \mathbb{R}$ denotes the root joint's angular velocity along Y-axis, linear velocity along X-axis and Z-axis, respectively. $r^h \in \mathbb{R}$ is the height of the root joint. $\mathbf{j}^p, \mathbf{j}^v \in \mathbb{R}^{J \times 3}$ are the position and linear velocity of each joint, where $J$ is the number of joints. $\mathbf{j}^r \in \mathbb{R}^{J \times 6}$ is the 6D rotation~\citep{zhou2019continuity} of each joint. Specifically, $J$ is 22 in HumanML3D and 21 in KIT-ML.


\paragraph{Quantitative Results.}\label{sec4_1_4}

We compare our proposed \name with five baseline models:  Language2Pose~\citep{ahuja2019language2pose}, Text2Gesture~\citep{bhattacharya2021text2gestures}, MoCoGAN~\citep{tulyakov2018mocogan}, Dance2Music~\citep{lee2019dancing}, and \cite{guo2022generating}. All baselines' performances are quoted from \cite{guo2022generating}. Table~\ref{tab:humanml3d} and Table~\ref{tab:kit} show the quantitative comparison on the HumanML3D dataset and the KIT-ML dataset. Our proposed \name outperforms all existing works with a remarkable margin in aspects of precision, FID, MultiDodal Distance, and Diversity. The precision of \name is even close to that of real motions, which suggests that our generated motion sequences are satisfyingly high-fidelity and realistic. 

~\cite{guo2022generating} states that the results on the MultiModality metric should be larger whenever possible. However, the literature in action-conditioned motion generation task~\citep{guo2020action2motion,petrovich2021action,cervantes2022implicit} argue that this metric should be close to the real motion. In the T2M task, it is difficult to calculate this metric of real motions. Therefore, we only report these results without comparison.



\begin{table}[ht]
\centering
\caption{\textbf{Ablation of the pretrained CLIP and the efficient attention technique.} All results are reported on the KIT-ML test set.}
\label{tab:kit_ablation}
\setlength{\tabcolsep}{1.4mm}
{
\begin{tabular}{ccccc}
\hline

\multirow{2}{1cm}{CLIP} & \multirow{2}{1cm}{\centering EFF} & \multicolumn{3}{c}{\centering R Precision$\uparrow$}  \\
& & Top 1 & Top 2 & Top 3 \\
\hline
N & N & $0.288^{\pm.004}$ & $0.440^{\pm.004}$ & $0.539^{\pm.004}$ \\
N & Y & $0.136^{\pm.003}$ & $0.233^{\pm.003}$ & $0.309^{\pm.003}$ \\
Y & N & $0.357^{\pm.004}$ & $0.555^{\pm.004}$ & $0.679^{\pm.005}$ \\
Y & Y & $\mathbf{0.417^{\pm.004}}$ & $\mathbf{0.621^{\pm.004}}$ & $\mathbf{0.739^{\pm.004}}$ \\
\hline
\end{tabular}}
\end{table}

To further understand the function of CLIP initialization and efficient attention, we report ablation results in Table ~\ref{tab:kit_ablation}. The models without pretrained CLIP suffer from severe performance drops, which indicates the necessity of a pretrained language model for the T2M task. As for efficient attention, it is significantly beneficial when we use CLIP simultaneously. However, this module also limits the model's performance when without CLIP. A possible explanation for this phenomenon is that the global relation in efficient attention is misleading when the semantic information from given text is insufficient.

\begin{table}[ht]
\centering
\caption{\textbf{Ablation of the latent dimension and the number of transformer layers.} All results are reported on the KIT-ML test set.}
\label{tab:kit_feat_layer}
\setlength{\tabcolsep}{1.4mm}
{
\begin{tabular}{ccccc}
\hline

\multirow{2}{1cm}{\centering \#layers} & \multirow{2}{1cm}{\centering Dim} & \multicolumn{3}{c}{\centering R Precision$\uparrow$}  \\
& & Top 1 & Top 2 & Top 3 \\
\hline
4 & 128 & $0.033^{\pm.002}$ & $0.066^{\pm.003}$ & $0.097^{\pm.003}$ \\
4 & 256 & $0.095^{\pm.002}$ & $0.166^{\pm.003}$ & $0.227^{\pm.003}$ \\
4 & 512 & $0.405^{\pm.005}$ & $0.620^{\pm.005}$ & $\mathbf{0.743^{\pm.004}}$ \\
8 & 128 & $0.025^{\pm.002}$ & $0.053^{\pm.002}$ & $0.086^{\pm.002}$ \\
8 & 256 & $0.198^{\pm.003}$ & $0.335^{\pm.004}$ & $0.441^{\pm.004}$ \\
8 & 512 & $\mathbf{0.417^{\pm.004}}$ & $\mathbf{0.621^{\pm.004}}$ & $0.739^{\pm.004}$ \\
12 & 128 & $0.031^{\pm.002}$ & $0.063^{\pm.003}$ & $0.091^{\pm.002}$ \\
12 & 256 & $0.209^{\pm.003}$ & $0.348^{\pm.004}$ & $0.452^{\pm.003}$ \\
12 & 512 & $0.412^{\pm.006}$ & $0.616^{\pm.004}$ & $0.741^{\pm.004}$ \\
\hline
\end{tabular}}
\end{table}

We explore how the size of architecture influences the performance. Table ~\ref{tab:kit_feat_layer} suggests that the latent dimension plays a more important role. The models with 512 latent dimension significantly outperform the models with 256 latent dimension. On the contrary, the increase of the number of layers improves the performance when the latent dimension is either 128 or 256, but has little effect when the dimension is 512.

\paragraph{Qualitative Results}\label{sec4_1_4}

\begin{figure*}[t]
    \centering
    \includegraphics[width=\linewidth]{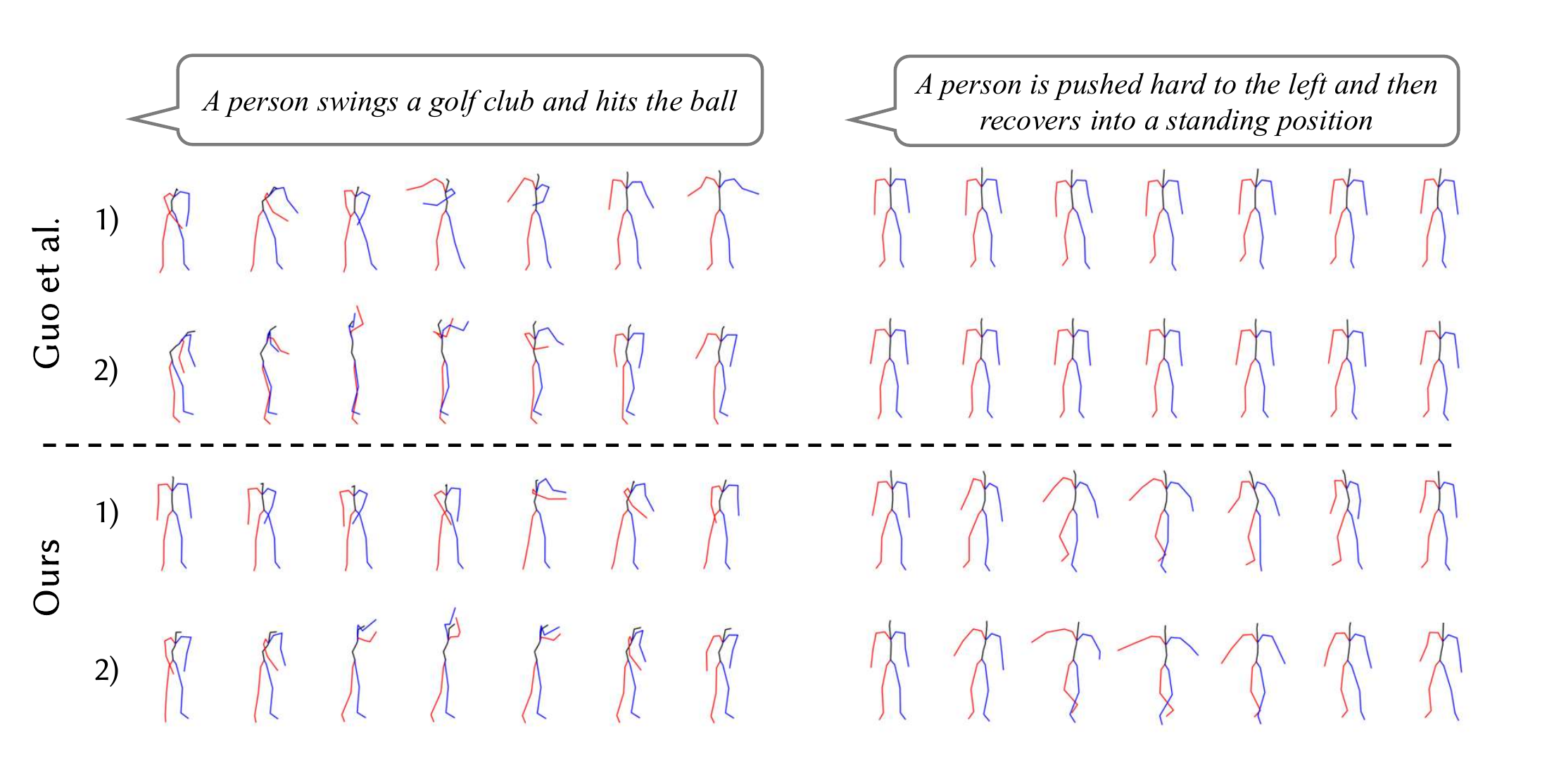}
    \caption{\textbf{Qualitative results on the HumanML3D dataset.} We compare our method with \cite{guo2022generating} and visualize two examples for each given prompt. \name is able to achieve both accuracy and diversity.}
    \label{fig:sec4_1}
\end{figure*}

Figure~\ref{fig:sec4_1} shows a comparison between our method and \cite{guo2022generating} as a baseline. We highlight that \name achieves a balance between diversity and realness. For example, for prompt `A person swings a golf club and hits the ball', our generated motions portraits the described motion more faithfully. In contrast, the baseline method has high multi-modality at the expenses of accuracy. In addition, given a complicated prompt such as "A person is pushed hard to the left and then recovers into a standing position", \name is able to generate high-quality motions that reflects the detailed description whereas the baseline method fails to produce any meaningful movement.

\subsection{Action-conditioned Motion Generation}\label{sec4_2}

\begin{table*}[ht]
\centering
\caption{\textbf{Quantitative results for Action-conditioned Motion Generation.} As for UESTC dataset, we report FID on the test split. MM: MultiModality.}
\label{tab:action}
\setlength{\tabcolsep}{1.4mm}
{
\begin{tabular}{lcccccccc}
\hline

\multirow{2}{2cm}{\centering Methods} & \multicolumn{4}{c}{\centering HumanAct12} & \multicolumn{4}{c}{\centering UESTC} \\

& FID$\downarrow$ & Accuracy$\uparrow$ & Diversity$\rightarrow$ &  MM$\rightarrow$ & FID$\downarrow$ & Accuracy$\uparrow$ &  Diversity$\rightarrow$ &  MM$\rightarrow$ \\

\hline
Real motions & $0.020^{\pm.010}$ & $0.997^{\pm.001}$ & $6.850^{\pm.050}$ & $2.450^{\pm.040}$ & $2.79^{\pm.29}$ & $0.988^{\pm.001}$ & $33.34^{\pm.320}$ & $14.16^{\pm.06}$\\ 
\hline
Action2Motion & $0.338^{\pm.015}$ & $0.917^{\pm.003}$ & $6.879^{\pm.066}$ & $2.511^{\pm.023}$ & - & - & - & - \\
ACTOR & $0.12^{\pm.00}$ & $0.955^{\pm.008}$ & $6.84^{\pm.03}$ & $2.53^{\pm.02}$ & $23.43^{\pm2.20}$ & $0.911^{\pm.003}$ & $31.96^{\pm .33}$ & $\mathbf{14.52^{\pm.09}}$ \\
INR & $0.088^{\pm.004}$ & $0.973^{\pm.001}$ & $6.881^{\pm.048}$ & $2.569^{\pm.040}$ & $15.00^{\pm.09}$ & $0.941^{\pm.001}$ & $31.59^{\pm.19}$ & $14.68^{\pm.07}$ \\

\hline
Ours & $\mathbf{0.07^{\pm.00}}$ & $\mathbf{0.992^{\pm.13}}$ & $\mathbf{6.85^{\pm.02}}$ & $\mathbf{2.46^{\pm.02}}$ & $\mathbf{9.10^{\pm.437}}$ & $\mathbf{0.950^{\pm.000}}$ & $\mathbf{32.42^{\pm.214}}$ & $14.74^{\pm.07}$ \\
\hline
\end{tabular}}
\end{table*}

\paragraph{Datasets.}\label{sec4_2_1} HumanAct12 dataset~\citep{guo2020action2motion} provides 12 kinds of motion sequences. This dataset is adapted from PHSPD dataset~\citep{zou20203d}, which contains 1191 videos. HumanAct12 further arranges these videos into trimmed motion clips. UESTC dataset~\citep{ji2018large} is also a significant benchmark for action-conditioned motion generation tasks, which includes 25K motion sequences across 40 different action categories. ~\cite{petrovich2021action} further uses pre-trained VIBE~\citep{kocabas2020vibe} to extract SMPL~\citep{loper2015smpl} sequences from the UESTC dataset and provides pretrained action recognition model for evaluation.

\paragraph{Evaluation Metrics.}\label{sec4_2_2} Four evaluation metrics are applied for this task: FID, Accuracy, Diversity, and Multimodality. The pretrained action recognition module can directly calculate the average accuracy for all action categories without arranging mini-batches. This metric has a similar function to R Precision. The other three metrics have been introduced in Section ~\ref{sec4_1}. HumanAct12 has no official split, and we report the FID on the whole dataset. UESTC has a test split, so we report the FID on it, which is more representative than the train split. In this section, FID and Accuracy are two principal metrics. Our conclusion are mainly based on them.

\paragraph{Implementation Details.}\label{sec4_2_3}
All the setting are the same to those for text-driven motion generation tasks except for the learning rate, the number of iterations and the motion representation. In this series of expreiments, we train 100K iterations for the HumanAct12 dataset and 500K for the UESTC dataset, both with a 0.0001 learning rate.

Motion represention in this task is slightly different from the T2M task. As for the HumanAct12 dataset, each pose state can be representated as $(\mathbf{j}^x, \mathbf{j}^y, \mathbf{j}^z)$, where $\mathbf{j}^x, \mathbf{j}^y, \mathbf{j}^z \in \mathbb{R}^{24 \times 3}$ are the coordinates of 24 joints. We use $(r^x, r^y, r^z, \mathbf{j}^r)$ as the pose representation for the UESTC dataset, where $r^x, r^y, r^z \in \mathbb{R}$ are the coordinates of the root joint, and $\mathbf{j}^r \in \mathbb{R}^{24 \times 6}$ is the rotation angle of each joint in 6D representation.


\paragraph{Quantitative Results.}\label{sec4_2_4} Following ~\cite{cervantes2022implicit}, three baseline models are selected: Action2Motion~\citep{guo2020action2motion}, ACTOR~\citep{petrovich2021action}, INR~\citep{cervantes2022implicit}. Table ~\ref{tab:action} shows the quantitative results on the HumanAct12 dataset and the UESTC datasets. Our proposed \name achieves the best performance in aspects of FID and Accuracy when compared to other existing works. We want to highlight that our results of the HumanAct12 dataset are notably close to real motions on all four metrics.

\begin{table}[ht]
\centering
\caption{\textbf{Ablation of the latent dimension and the number of transformer layers.} All results are reported on the HumanAct12 dataset.}
\label{tab:action_feat_layer}
\setlength{\tabcolsep}{1.4mm}
{
\begin{tabular}{cccc}
\hline

\#layers & Dim & FID$\downarrow$ & Accuracy$\uparrow$  \\
\hline
4 & 128 & $0.29^{\pm0.00}$ & $0.892^{\pm1.97}$ \\
4 & 256 & $0.14^{\pm0.00}$ & $0.958^{\pm0.51}$ \\
4 & 512 & $0.09^{\pm0.00}$ & $0.984^{\pm0.21}$ \\
8 & 128 & $0.22^{\pm0.00}$ & $0.929^{\pm1.04}$ \\
8 & 256 & $0.09^{\pm0.00}$ & $0.983^{\pm0.23}$ \\
8 & 512 & $\mathbf{0.07^{\pm0.00}}$ & $0.992^{\pm0.13}$ \\
12 & 128 & $0.11^{\pm0.00}$ & $0.954^{\pm0.67}$ \\
12 & 256 & $0.10^{\pm0.00}$ & $0.988^{\pm0.21}$ \\
12 & 512 & $0.08^{\pm0.00}$ & $\mathbf{0.996^{\pm0.08}}$ \\
\hline
\end{tabular}}
\end{table}

Here we also try different combination of latent dimension and the number of layers, as shown in Table ~\ref{tab:action_feat_layer}. Similar to the conclusions we found in Section~\ref{sec4_1}, latent dimension is more important than the number of layers.

\subsection{Motion Manipulation}\label{sec4_3}

To better evaluate the capability of text-driven motion generation models, we design two task variants. 
First, \textit{Spatially-diverse T2M task (T2M-S)}. T2M-S requires the generated motion sequence to contain multiple actions on different body parts (\eg `a person is running and drinking water simultaneously'). Specifically, $i$-th test sample in T2M-S task can be represented by a set of text-mask pairs $\{(\textrm{text}_{i,j}, \mathrm{M}_{i,j})\}$, where $\mathrm{M}_{i,j} \in \{0,1\}^D$ is a $D$-dimension binary vector. It indicates which body part we should focus on when given the text description $\textrm{text}_{i,j}$.
Second, \textit{Temporally-diverse T2M task (T2M-T)}. T2M-T expects models to generate a long motion sequence, which includes multiple actions in a specific order spanning over different time intervals (\eg `a person is walking and then running'). The $i$-th test sample is an array of text-duration pairs $\{\textrm{text}_{i,j}, [l_{i,j}, r_{i,j}]\}, l_{i,j} < r_{i,j}$. It means that the motion clip from  $l_{i,j}$-th frame to $r_{i,j}$ frame is supposed to contain the action $\textrm{text}_{i,j}$.

\paragraph{Implementation Details.}\label{sec4_3_2} 
We train our proposed \name on the BABEL dataset~\citep{punnakkal2021babel} with 50K iterations. Each pose state is represented by $(r^x, r^y, r^z, \mathbf{j}^r)$, which is same to the setting for the UESTC dataset. Other settings remain unchanged. $\lambda_1 =\lambda_2 = 0.01$ are used for the visualization.

\begin{figure*}[t]
    \centering
    \includegraphics[width=\linewidth]{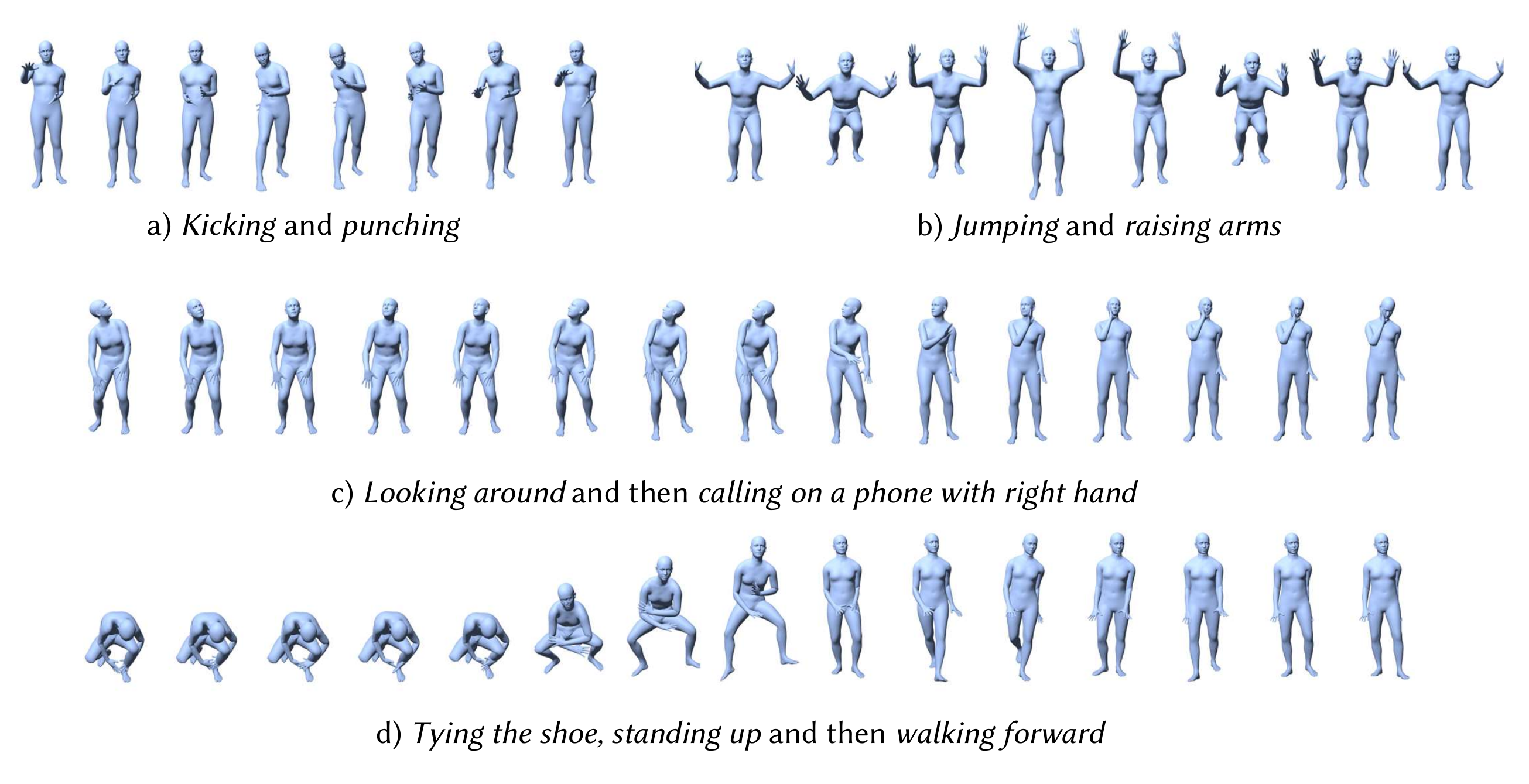}
    \caption{\textbf{Qualitative results on the BABEL dataset.} \name is able to generate dynamic sequences according to complicated prompt that involves multiple body parts or actions.}
    \label{fig:sec4_3}
\end{figure*}

\paragraph{Qualitative Results.}\label{sec4_3_4}
As shown in Fig \ref{fig:sec4_3}, \name has the capability to handle highly comprehensive prompts that assign motions to multiple body parts (such as ``Kicking and punching" and ``Jumping and raising arms" that require coordination of the upper and lower body). Moreover, \name is able to generate long sequences according to a complex instruction that includes multiple actions (such as ``Tying the shoe, standing up and then walking forward" that includes a series of vastly different motions).

\section{Conclusion, Limitations and Future Work}\label{sec5}
We propose \name, the first diffusion model-based method for text-driven motion generation. \name demonstrates three major strengths: Probabilistic Mapping that enhances diversity, Realistic Synthesis that ensures plausibility of motion sequences, and Multi-Level Manipulation that allows for per-part manipulation and long sequence generation. Both quantitative and qualitative evaluations show that \name outperforms existing arts on various tasks such as text-driven motion generation and action-conditioned motion generation, and demonstrates remarkable motion manipulation capabilities.

Although \name has pushed forward the performance boundary of motion generation tasks, there still exist some problems. First, diffusion models require a large amount of diffusion steps during inference and it is challenging to generate motion sequences in real-time. Second, current pipeline only accepts a single form of motion representation. A more generalized pipeline that adapts concurrently to all datasets would be more versatile for various scenarios.

\begin{acknowledgements}
This work is supported by NTU NAP, MOE AcRF Tier 2
(T2EP20221-0033), and under the RIE2020 Industry Alignment Fund - Industry Collaboration Projects (IAF-ICP) Funding Initiative, as well as cash and in-kind contribution from the industry partner(s).
Corresponding author: Ziwei Liu (ziwei.liu@ntu.edu.sg).
\end{acknowledgements}

%
%

\bibliographystyle{spbasic}      
\bibliography{ref}   


\end{document}